\documentclass[runningheads,a4paper]{llncs}

\usepackage{amssymb,amsmath}
\usepackage{graphicx}

\usepackage{subfigure, caption}

\usepackage{url}

\begin{document}

\mainmatter  

\title{A Denoising Autoencoder that Guides Stochastic Search}

\titlerunning{A Denoising Autoencoder that Guides Stochastic Search}

%
%
\author{Alexander W. Churchill %
\and Siddharth Sigtia \and Chrisantha Fernando}
\authorrunning{Alexander W. Churchill, Siddharth Sigtia,  and Chrisantha Fernando}

\institute{School of Electronic Engineering and Computer Science\\Queen Mary, University of London\\
{\{a.churchill,s.s.sigtia,c.t.fernando\}}@qmul.ac.uk}

%
%

\toctitle{Lecture Notes in Computer Science}
\tocauthor{Authors' Instructions}
\maketitle

\begin{abstract}
An algorithm is described that adaptively learns a non-linear mutation distribution. It works by training a denoising autoencoder (DA) online at each generation of a genetic algorithm to reconstruct a slowly decaying memory of the best genotypes so far. A compressed hidden layer forces the autoencoder to learn hidden features in the training set that can be used to accelerate search on novel problems with similar structure. Its output neurons define a probability distribution that we sample from to produce offspring solutions. The algorithm outperforms a canonical genetic algorithm on several combinatorial optimisation problems, e.g. multidimensional 0/1 knapsack problem, MAXSAT, HIFF, and on parameter optimisation problems, e.g. Rastrigin and Rosenbrock functions.
\end{abstract}

\section{Introduction}
Evolutionary algorithms do not use supervised learning methods to learn to predict which mutants are likely to be fitter than others, perhaps because natural selection in the wild, their original biological inspiration \cite{Holland1975}, does not exhibit such mechanisms. However, the theory of {\em{evolutionary neurodynamics}} \cite{fernando2010neuronal} proposes that evolutionary computation takes place in the brain, and that it can exploit Hebbian learning for learning linkage between alleles at distant loci \cite{fernando2010neuronal}. This paper extends that work to learning more complex models than Hebbian learning alone is capable of. Our motivation is similar to the inventors of estimation of distribution algorithms (EDAs) which were developed to statistically model genotype space to bias exploration towards optimal solutions \cite{pelikan2006scalable}. Early work on EDAs concentrated on methods that explicitly modelled allele frequencies in the population \cite{harik1999compact,baluja1994population,pelikan2002survey}. Later, correlations between alleles at different loci were learned \cite{harik1999linkage}, and recently  Bayesian generative models have been learned, e.g. hBOA \cite{hboa}, albeit at considerable computational cost \cite{pelikan2002survey}. However, an underrepresented area in EDAs is transfer learning, i.e. generalising knowledge across different tasks, which is only beginning to be addressed, e.g.  in \cite{hauschild2012using}.

We combine a denoising autoencoder with a genetic algorithm to allow transfer learning \cite{caruana1997multitask}. An autoencoder is a feed forward neural network, consisting of at least one hidden layer, which is trained to reproduce its inputs from its outputs. Over the course of training, the hidden layer forms a compressed representation of the inputs \cite{hinton2006reducing}. Neural network methods have been employed before in this context, e.g. competitive Hebbian learning \cite{marti2008introducing}, restricted Boltzmann machines \cite{tang2010restricted} and Helmholtz machines \cite{zhang2000bayesian}. However, the DA strikes a desirable balance between learning a rich non-linear model of genotype space, rapidity, and biological plausibility \cite{caruana1997multitask,hinton2006reducing}.

\section{Methods}

\subsection{Denoising Autoencoder}

A standard autoencoder consists of an encoder and a decoder. The encoder performs an affine transformation followed by an element-wise non-linear operation. The mapping performed by the encoder is deterministic and can be described as  $h_{\theta}(\mathbf{x}) = f(\mathbf{Wx + b})$. The decoder is also a deterministic network and operates on the encoded input $h_{\theta}(x)$ to produce the reconstructed input $ r_{\theta'}(\mathbf{h}) = g(\mathbf{W'h + b'})$. The autoencoder, which is trained using Backpropagation, learns the distribution of the training data by means of the layer of hidden variables. The encoder network can be viewed as a non-linear generalisation of PCA \cite{hinton2006reducing}. The autoencoder learns interactions between the input attributes and maps it to the hidden layer via a non-linear transformation, making it a powerful model for learning and exploiting structure present in the best individuals.

We interpret the outputs of the decoder network as parameters for a conditional distribution $p(X|Z=\mathbf{z})$ over the outputs of the network given an input $\mathbf{z}$. For binary optimization problems, the outputs $\mathbf{z}$ are considered to be parameters for the distribution $X|\mathbf{z} \sim \mathcal{B(\mathbf{z}})$ where $\mathcal{B}$ is the Bernouilli distribution. For continuous parameter-optimisation problems, the outputs $\mathbf z$ parameterise a multi-variate normal distribution $X\ \sim\ \mathcal{N}_k(\mathbf{z},\,\sigma^2 \mathbf I)$, where the covariance matrix is assumed to be diagonal and the standard deviation along the diagonal is a tuneable parameter.

In our algorithm we use a denoising autoencoder which is a variant of the standard autoencoder \cite{vincent2008extracting}. The DA tries to recreate the input $\mathbf x$ from \em{corrupted}\em{} or \em{noisy}\em{} versions of the input, which are generated by a stochastic corruption process $\mathbf{\tilde x = q(\tilde x|x)}$. Autoencoders are trained along with strong regularisation in order to impose an information bottleneck on the model. If this is not done carefully, it would be very easy for the autoencoder to learn the identity transformation. Adding the denoising criterion to the model forces the autoencoder to learn transformations invariant to perturbations of the input. In our system we employ the denoising criterion to widen the basins of attraction around the best individuals in every generation. During training, we train the autoencoder on the most promising solutions, and by increasing corruption we encourage individuals that are far away from the training set to move towards the nearest high quality solution. By making use of the corruption noise as a tuneable parameter, the extent of the basins of attraction can be controlled, with high corruption giving rise to large basins of attraction.
 
\begin{figure}[t!]
\center
\hbox{
\hspace{-0.5em}
\includegraphics[width = 1.0\textwidth]{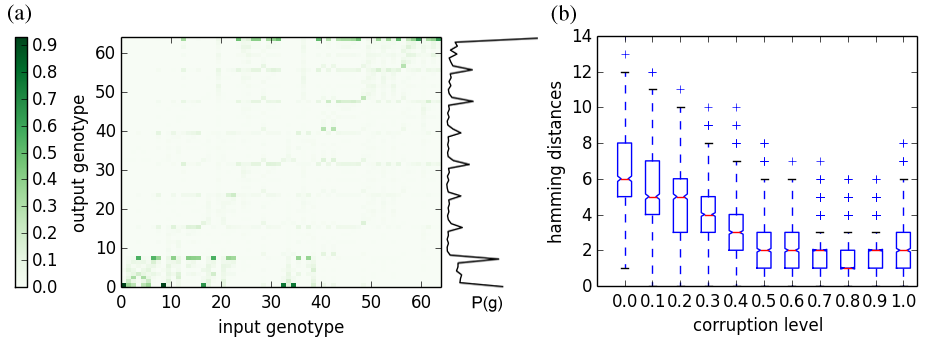}
}
\vspace{-10px}
\caption{Showing (a) transition probabilities (from inputs to outputs) for all possible genomes on the 6-bit problem, with the plot on the right showing probabilities of different outputs, \(P(g)\), marginalised over all possible inputs and (b) Hamming distances of a randomly sampled population from the optimal solution on a 20-bit MaxOnes problem with different corruption levels.} 
\label{figure:sids_graph}
\end{figure}

To demonstrate structure learning and the capacity of the DA to learn multiple examples, we train an autoencoder on 3 different 6-bit the target vectors: all ones, all zeros, and the 3 least significant bits set to 1. Fig~\ref{figure:sids_graph}(a) shows the matrix of transition probabilities for all possible pairs of 64 genotypes. The indices of the $X$ and $Y$ axes are labelled as the sequence of binary numbers: the 0-index corresponds to the string of all 0s and the 63rd index corresponds to the string of all ones. Each point (a,b) in the figure denotes the probability of transition from genome a to b. From the figure we observe that there are bands of high probabilities along $Y=0,7,63$. This implies that for most input genomes, the autoencoder has a high probability of producing a genome that is close to one of the target solutions. In Fig~\ref{figure:sids_graph}(a), we also plot the marginalised probabilities of each of the target genomes. We observe peaks at the location of the three target vectors. We also notice that there is a band of high probability between the vectors 000000 and 000111. This demonstrates that the autoencoder learns the structure of the best solutions and yields outputs that are closer to the best solutions. The fact that the new population has probability peaks around all three target solutions demonstrates that the autoencoder learns a complex non-linear mapping from input to output and is capable of retaining information about several classes of solutions.   

To demonstrate the influence of corruption noise on widening the basin of attraction around the training examples we train the GA on a 20-bit MaxOnes problem. MaxOnes has a unique optimum solution and the Hamming distances from the optimal solution are easy to interpret. We allow the GA to run for 10 generations, with a population of 200 and train the autoencoder on the top 20 solutions at every generation. The inputs to the DA are corrupted by a binary corruption process that stochastically flips a fixed proportion of bits for each example. The DA is then trained to reconstruct the true targets from the corrupted inputs. After training, 100 randomly generated input vectors are presented to the DA, and a new population is sampled. We repeat this experiment with different corruption rates and the same random input vectors and plot the Hamming distances of the sampled population from the optimum solution. Fig~\ref{figure:sids_graph}(b) shows that increasing the corruption rate leads to a decrease in the Hamming distances of offspring from the target. This shows that as corruption noise is increased, the outputs of the DA are more likely to be closer to an example in the training set.
\vspace{-15px}
\subsection{Optimisation Algorithm}

The pipeline of the \emph{Denoising Autoencoder Genetic Algorithm} (DAGA) is similar to EDAs and is inspired by methods in HBOA \cite{hboa}. A population of solutions is maintained, \(P_t\), and updated at each iteration, \(t\). An initial population of solutions, \(P_0\), is drawn from a uniform distribution. These solutions are evaluated and the fittest unique x\% are selected (i.e. truncation selection) to be in the training set, \(\hat{P_t}\). The Denoising Autoencoder, \(D_t\), is then trained with \(\hat{P_t}\) for \(e\) epochs. Following training, a new set of solutions, \(S_t\), is selected from \(P\) using tournament selection (with replacement). Each member of \(S_t\) is inputted to \(D_t\), and the output vector, \(y\), is sampled from using a binomial distribution, to produce a new solution. This solution is included in \(P_{t+1}\) if it is better than its closest neighbour according to Restricted Tournament Selection (RTR) \cite{hboa}.
\section{Experiments}
\label{sec:experiments}
DAGA is tested on three sets of experiments: (1) discrete, (2) continuous and (3) accumulation of adaptation problems. We perform experiments on two different instances of the Multi-dimensional Knacksack problem. The first is the Weing8 instance \cite{weingartner1967methods}, which has 105 items and two constraints (optimal solution is 602,319), and the second is a randomly generated instance with 500 items and one constraint (optimal solution is 10,104). Both instances are available in the supplementary material.\footnote{\small{Online supplementary material: \small{\url{http://www.robozoo.co.uk/research/ppsn2014}}}}. The proposed system is then applied to the 128-bit and the 256-bit Hierarchical If and Only If problem (HIFF) \cite{watson1999hierarchically}. In order to remove any possible bias towards an all 1s solution, solutions from DAGA have a random bit mask (fixed before each trial) applied to them before evaluating fitness. We also test the system on a 128-bit Royal Road problem with 8-bit partitions \cite{mitchell1992royal}. As with the HIFF, a random mask is applied to the output of DAGA before fitness evaluation. Finally, the system is tested on a 3-CNF, 100-bit MAXSAT problem obtained from SATLIB \cite{hoos2000satllb}. The problem consists of 430 clauses and belongs to the phase transition region, which is the point at which the problem transitions from generally solvable to generally unsolvable. 

We evaluate our system's performance on continuous parameter optimisation problems by applying it to a 50-d Sphere, a 10-d Rosenbrock and a 10-d Rastrigin function. The Sphere function is given by $f(p) = \sum_{i=1}^{n} p_i^{2}, -5.12<p_{i}<5.12$. The Rosenbrock function is $f(p) = \sum_{i=1}^{n} 100(p_{i+1}-p_i^{2})^2 + (1+p_i)^2, -2.048 \le p_{i} \le 2.048$. And finally the Rastrigin function is defined as $f(p) = 10.n + \sum_{i=1}^n \left[p_i^2 - 10.\cos(2 \pi p_i)\right], -5.12 \le p_{i} \le 5.12$. 

We also investigate whether DAGA can be used for transfer learning, i.e. to achieve faster convergence on new tasks, that are similar to tasks that it has already learnt to solve. In order to test the validity of this claim, a new task was devised that was influenced by Watson et al \cite{watson2014evolution}. DAGA is applied to solve three problem instances in sequence, without reinitialising the weights of the DA between instances. The problems consist of minimising the Hamming distance to a target string, which represents an 81-pixel image. The algorithm is evaluated on the speed up created by solving the third problem, where speed up is defined as the difference in the number of iterations needed to solve the third problem after solving the first two, compared to solving the third problem ab initio. The first and second solutions are a Cross and Box pattern respectively. The third problem is a pattern similar to the original patterns. DAGA is trained on the Cross and the Box (XB) and then on one of the 16 similar patterns, as well as the Box and the Cross (BX) and then on one of the other combined patterns, resulting in 32 transfer learning experiments. 

\section{Results}
\label{section:results}
On the discrete and continuous problems, DAGA is compared to a canonical generational Genetic Algorithm (GA). On the continuous problems it is additionally compared to a (1+1)-ES with the 1/5 rule \cite{beyer2002evolution}. The GA employs tournament selection, two point crossover (with probability \(p_c\)) and on the discrete problems there is a probability, \(p_m\), of a bit flip mutation at each allele, while on the continuous problems there is probability, \(p_m\), of adding Gaussian noise to each component of the vector from \(N(0, \delta^2)\). Programming code is available in the online supplementary material, as well as fitness graphs for each experiment. For each problem a large parameter sweep was performed on both DAGA and the GA and the best configurations (averaged over 10 samples per parameter set) were chosen for comparison. 

Table~\ref{table:main_results} presents details of the best solution returned at the end of a run and the number of evaluations required for DAGA and the GA on the 6 problems described above in Section 3. The table shows that DAGA finds either significantly better solutions, or optimal solutions significantly faster than the GA, on all of the problems apart from the Royal Road where they are not significantly different. On four of the six experiments (both knapsacks, MAXSAT and 256-bit HIFF), DAGA finds significantly better solutions than the GA, within the given evaluation limits. On the Weing8 knapsack instance DAGA reaches the optimum solution in 8 out of 10 attempts, and on the 500-item knapsack in 7 out of 10 attempts. On the MAXSAT problem, DAGA reaches the optimum 70\% of the time. On all three of these problems the GA is unable to find each optimal solution once. On the 128-bit HIFF, DAGA locates the optimal solution every time, while the GA locates it on half of the trials, and on the 256-bit HIFF DAGA again has a 100\% success rate while the GA achieves only 30\%. On the Royal Road, DAGA locates the global optima on every trial, while the GA finds it on 9 out of 10. These results show that on a wide range of discrete problems DAGA is able to find higher quality solutions and more consistently locate the optimum compared to a GA. As well as frequently obtaining better quality solutions at the end of the optimisation process, DAGA also finds optimal solutions with fewer evaluations than the GA. On all problems, Table~\ref{table:main_results} shows that DAGA has a lower number of average evaluations needed to find the optimum, with statistical significance on MAXSAT, 128-HIFF and Knapsack 500.
 \begin{table}[t!]
   \scalebox{0.9}{
    \begin{tabular}{ | p{1.8cm} | l | l | l | l | l | l | l | l | p{1cm} |}
    \hline
    Experiment & Algorithm & Min & Max & Mean & Mean Evals & Success \%\\ \hline
MAXSAT & GA & 424 & 429 & 426.5 \(\pm\)1.4 & 500000 \(\pm\)0.0 & 0\%\\
& AE & 429 & 430 & \textbf{429.8 \(\pm\)0.4\(^*\)} & \textbf{291944.4 \(\pm\)134968\(^*\)} & 70\%\\\hline
128 HIFF & GA & 832 & 1024 & 940.8 \(\pm\)86.1 & 416000 \(\pm\)87430 & 50\%\\
& AE & 1024 & 1024 & 1024 \(\pm\)0.0 & \textbf{231000 \(\pm\)23537.2\(^*\)} & 100\%\\\hline
256 HIFF & GA & 1664 & 2304 & 1984 \(\pm\)225.4 & 1590500 \(\pm\)626719.4 & 30\%\\
& AE & 2304 & 2304 & \textbf{2304 \(\pm\)0.0\(^*\)} & 1355500 \(\pm\)190793.7 & 100\%\\\hline
Knapsack Weing8 & GA & 619568 & 621086 & 620099.2 \(\pm\)529.8 & 100000 \(\pm\)0.0 & 0\%\\
& AE & 621086 & 624319 & \textbf{623124.3 \(\pm\)1416.6\(^*\)} & 96600 \(\pm\)34414.2 & 80\%\\\hline
Knapsack 500 & GA & 10047 & 10093 & 10074.2 \(\pm\)14.0 & 200000 \(\pm\)0.0 & 0\%\\
& AE & 10096 & 10104 & \textbf{10102.7 \(\pm\)2.5\(^*\)} & \textbf{121620 \(\pm\)52114.9\(^*\)} & 70\%\\\hline
Royal Road & GA & 120 & 128 & 127.2 \(\pm\)2.4 & 42600 \(\pm\)27034.9 & 90\%\\
& AE & 128 & 128 & 128 \(\pm\)0.0 & 26500 \(\pm\)3721.6 & 100\%\\\hline
    \end{tabular}
    }
    \caption{Results for DAGA and the GA on 6 discrete problems. Showing the value of the minimum and maximum solution returned at the end of search, the mean best solution in the population after the end of search, the mean number of evaluations required to reach the optimum solution (or until the maximum number of evaluations) and the success rate for reaching the optimum, averaged across 10 trials. Stars indicate that the mean result for the specified algorithm is significantly different from the GA according to a Wilcoxan Rank Sum test (\(p<0.05\)).}
    \vspace{-20px}
        \label{table:main_results}

\end{table}

Results for DAGA, (1+1)-ES and the GA are presented in Table~\ref{table:continuous_results} for the Sphere, Rosenbrock and Rastrigin functions. On all three of these problems DAGA outperforms the GA, and is able to reach significantly better solutions with fewer evaluations. However, the (1+1)-ES is considerably better on the Sphere and Rosenbrock functions, achieving significantly better solutions than DAGA over the whole run and in far fewer evaluations. On the Rosenbrock it is almost three times as fast, and on the Sphere can reach the 0.1 target after only 235 evaluations, more than 100 times faster than DAGA. This implies that the (1+1)-ES is a much more efficient continuous optimiser on these search spaces. However, the (1+1)-ES performs extremely badly on the Rastrigin, which consists of many local optima. Here, the population-based GA and DAGA excel, getting much closer to the optimal solution of 0. DAGA is significantly better than the GA and the only algorithm to find solutions under 1.0.
 \begin{table}[t!]
 \scalebox{0.94}{
    \begin{tabular}{| l | l | l | l | l | l | l |}
    \hline
    Experiment & Algorithm & Target & Min & Max & Mean & Mean evals  \\ \hline
Sphere 50D & GA & 0.1 & 0.065 & 0.101 & 0.088 \(\pm\)0.016 & 49300\(\pm4368.829\)\\
 & (1+1)-ES & 0.1 & 0.000 & 0.000 & \textbf{0.00 \(\pm\)0.000} & 235\(\pm23.240\)\\
 & dAGA & 0.1 & 0.026 & 0.033 & 0.029 \(\pm\)0.003 & 36150\(\pm502.494\)\\\hline
 Rosenbrock 10D & GA & 0.1 & 0.547 & 0.800 & 0.659 \(\pm\)0.106 & 300000\(\pm0\)\\
 & (1+1)-ES & 0.1 & 0.000 & 0.000 & \textbf{0.000 \(\pm\)0.000} & 27192.2\(\pm1186.672\)\\
 & dAGA & 0.1 & 0.018 & 0.082 & 0.042 \(\pm\)0.018 & 81950\(\pm17329.815\)\\\hline
Rastrigin 10D & GA & 1.0 & 1.395 & 4.347 & 3.058 \(\pm\)1.234 & 300000\(\pm0\)\\
 & (1+1)-ES & 1.0 & 37.808 & 160.187 & 89.844 \(\pm\)33.520 & 300000\(\pm0\)\\
 & dAGA & 1.0 & 0.601 & 1.277 & \textbf{0.888 \(\pm\)0.203} & 167800\(\pm123007.154\)\\\hline
    \end{tabular}
    }
    
    \caption{Results for DAGA, (1+1)-ES and the GA on 3 continuous problems. Showing the value of the minimum and maximum solution returned at the end of search, the mean best solution in the population after the end of search, the mean number of evaluations required to reach the target value (or until the maximum number of evaluations).  On each function results were found to have significant differences on a one way ANOVA (\(p<0.05\)), post-hoc analysis with a t-test and the Bonferroni correction for multiple comparisons showed significant differences between all pairs (\(p<0.05\)).}\vspace{-25px}
    \label{table:continuous_results}
\end{table}

\begin{figure}[t!]
\center
\hbox{
\hspace{-6.5em}
\includegraphics[width = 1.3\textwidth]{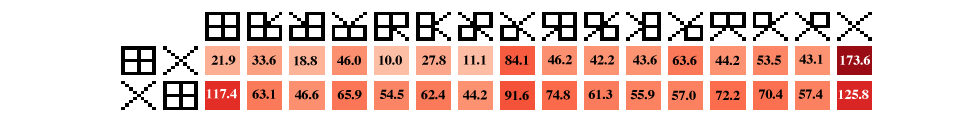}
}
\caption{Showing the average speed up on the transfer learning task, the difference between mean number of iterations taken to solve the image pictured on the top row without pre-learning and the mean number of iterations taken with pre-learning on the two solutions pictured in the first column, averaged over 10 trials. All mean iterations are significantly different, Wilcoxan Rank Sum test (\(p<0.05\))\vspace{-10px}} 
\label{figure:cross_square_grid}
\end{figure}

\begin{figure}[t!]
\center
\hbox{
\hspace{2.0em}
\includegraphics[width = 0.8\textwidth]{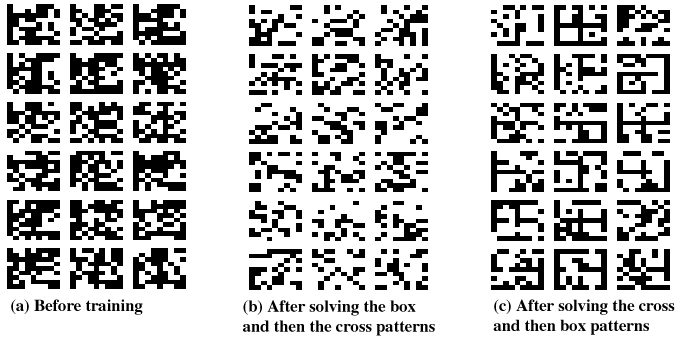}
}
\caption{Showing the output of 18 random inputs passed through the autoencoder (a) without training at the start of the optimisation process, (b) after solving the box and then the cross patterns (c) after solving the cross and then the box.} 
\label{figure:patterns}
\end{figure}

Results for DAGA on the image generation task described in Section 3 are presented in Fig~\ref{figure:cross_square_grid}. This shows the difference in the average number of iterations taken to solve the third problem when solving the third problem directly compared to when solving it using a DA that has been trained on solving the first two problems without reinitialisation. We see that in every case, there is a significant speed up when DAGA solves the first two problems in the sequence. Fig.~\ref{figure:patterns}(a) shows the outputs of the DA generated without training, Fig.~\ref{figure:patterns}(b) shows the population created by sampling random inputs after solving the Box and Cross patterns in sequence (this set will be referred to as BX), and Fig.~\ref{figure:patterns}(c) shows the same after solving the Cross and Box patterns (XB). There is a clear difference in the solutions created, with Fig.~\ref{figure:patterns}(a) showing much noisier solutions without much structure, Fig.~\ref{figure:patterns}(b) showing cross-like patterns mixed with block elements and Fig.~\ref{figure:patterns}(c) showing box patterns with diagonal elements. We can see that by training on two solutions, the DA has learned from both training patterns, and this is exploited to solve the third problem considerably faster, as seen in Fig. \ref{figure:cross_square_grid}.

Unsurprisingly, the largest speedups occur when the third problem is identical to the second. For both initial training patterns, this speedup is considerably greater than the mean, although it is markedly greater on the BX set. Also notable on BX is that the speed up is much lower on the Box pattern, even though it was already solved in the first run of the algorithm. By training on the Cross pattern second, some information learnt from the first run appears to have been lost. On BX, apart from the Cross, the largest speed up is seen in the solutions which inherit \(1/4\) from the Box and \(3/4\) from the Cross. The reverse is not true, and for XB there are large savings on the Cross, which it was trained on initially, as well as the \(1/4\) Box, \(3/4\) Cross solution that BX performs well on. Fig.~\ref{figure:patterns} (b) and (c) shows that while BX and XB both share features of the two root patterns, they are dominated by the second trained pattern. The Cross patterns are more sparse, which could explain why it takes longer to produce a Box, as more pixels must be filled in. In order to reduce the memory effect, it may help to use a lower learning rate. However, this simple example illustrates that DAGA is able to transfer learned structure between tasks.
 \begin{table}[t!]
  \scalebox{0.8}{
    \begin{tabular}{ | l | l | l | l | l | l | l | l | l | l | l | l | l |}
    \hline
    & \multicolumn{3}{|c|}{GA Parameters} & \multicolumn{7}{c|}{DAGA Parameters} \\\hline
    &Pop & CrossO & Mut & $\sigma^2$ & Pop & \% & Epochs & Learning Rate & Corruption & Hiddens & Niching\% & $\sigma^2$\\ \hline
    MAXSAT & 100 & 0.4& 0.01 & - & 2500 & 20 & 50 & 0.1 & 0.05 & 100 & 2500/20 & -\\
    128 HIFF & 10000 & 0.9 & $\frac{1}{10000}$ & - & 5000 & 10 & 25 & 0.1 & 0.05 & 128 & N/A & -\\
    256 HIFF & 15000 & 0.9 & $\frac{1}{15000}$ & - & 5000 & 20 & 50 & 0.01 & 0.05 & 500 & N/A & -\\
    Knapsack 105 & 300 & 0.5 & $\frac{1}{300}$ & - & 500 & 10 & 100 & 0.1 & 0.25 & 80 & 500/2 & -\\
    Knapsack 500 & 200 & 0.9 & $\frac{1}{200}$ & - & 600 & 10 & 50 & 0.1 & 0.9 & 200 & 600/20 & -\\
    Royal Road & 400 & 0.9 & $\frac{1}{400}$ & - & 1000 & 10 & 50 & 0.1 & 0.9 & 100 & N/A & -\\
    Sphere & 100 & 0.3 & $\frac{1}{100}$ & 0.1 & 500 & 10 & 100 & 0.1 & 0.1 & 100 & N/A & $10^{-5}$\\
    Rosenbrock & 100 & 0.2 & 1 & 0.01 & 500 & 10 & 100 & 0.1 & 0.1 & 30 & N/A & $10^{-5}$\\
    Rastrigin & 1000 & 0.2 & $\frac{2}{100}$  & 0.01& 1000 & 20 & 200 & 0.1 & 0.1 & 25 & N/A & $10^{-5}$\\ 
    Transfer Learning &-&-&-& - & 1000 &20&10&0.0001&0.1&70& N/A & -\\ \hline

    \end{tabular}
    }
\label{table:params}
    \caption{Parameters for the GA and DAGA for all experiments.}\vspace{-15px}
\end{table}
\vspace{-10px}
\section{Discussion}
\vspace{-10px}The results from Section~\ref{section:results} show that the DA can be applied to both discrete and continuous problems and regularly outperforms a GA, either in the quality of the final solution or in the number of evaluations used. The advantages over the GA are most clear on difficult combinatorial problems such as the 256-bit HIFF and MAXSAT, where the GA struggles to locate optimal solutions. On these problems, the best DAGA parameters found through a grid search displayed a very low corruption level (0.05 for MAXSAT and 256-HIFF) and large population sizes (2,500 and 5,000). As seen in Fig.~\ref{figure:sids_graph} the low corruption level maintains variance in the population, maximising the amount of information in the data pool and preventing premature convergence. Conversely, on problems such as the single constraint Knapsack and Royal Road, a large corruption level is used. This implies that on problems where there is more independence between input components, a high corruption level can more quickly direct search towards desired regions.

While DAGA outperformed the GA on all three parameter optimisation problems, it was outperformed by a (1+1)-ES on two. There are several improvements that could be made, one of which is the sampling method, which uses a fixed \(\sigma^2\). Using an adaptive step-size, such as \em{Cumulative path-length control}\em{} \cite{beyer2002evolution} used in population-based Evolutionary Strategies could have a positive effect on convergence. To solve the continuous problems, DAGA was found to need a large hidden layer, a factor of 2.5 or greater than the input size. Continuous search spaces are much larger than discrete ones and the DA will need a high capacity in order to effectively capture solution structure. This was also found on the 256-bit HIFF, which used 500 hidden neurons. While this could lead to overfitting on the solution set, this can be offset by the corruption applied to the inputs, which acts as a strong regulariser. There is the potential that results could be improved through deeper architectures for the DA, which could capture a hierarchical composition of features, which a shallow architecture cannot.  

The transfer learning task poses an interesting question that has not been answered by the EDA community, although recent work has made a start (e.g. \cite{hauschild2012using}) - can building a model to solve one task help improve performance on another? The results showed that using a DA, information learned in one task can be retained and recombined with that learnt in a second task to produce an across the board speed up on the third task in the sequence. This shows the potential for using a DA, but also potentially other modelling techniques from the optimisation community, such as Restricted Boltzmann Machines or Deep Belief Networks, for transfer learning between problems of the same class. Careful experimentation is needed to discover which types of problem can use machine learning techniques to transfer knowledge between unique instances and speed up optimisation as more examples are encountered.

\section{Conclusion}
We have presented a novel stochastic algorithm based on a denoising autoencoder. Training online, using the best found genotypes, the algorithm is able adaptively learn an exploration distribution, which guides it towards optimal solutions on difficult problems such as the 256-bit HIFF, regularly outperforming a genetic algorithm. In a simple experiment, we have shown that in addition to standard optimisation problems, the DA can learn to generalise across instances in the same problem class, displaying speed ups on unseen instances. 

\subsubsection*{Acknowledgments.} The work is funded by the FQEB Templeton grant ``Bayes, Darwin and Hebb", and the FP-7 FET OPEN Grant INSIGHT.

\bibliographystyle{splncs}
\vspace{-20px}
\bibliography{autoencoder_ppsn}
\vspace{-20px}

\end{document}